\documentclass[10pt,twocolumn,letterpaper]{article}

\usepackage{cvpr}
\usepackage{times}
\usepackage{epsfig}
\usepackage{graphicx}
\usepackage{amsmath}
\usepackage{amssymb}

\usepackage[pagebackref=true,breaklinks=true,letterpaper=true,colorlinks,bookmarks=false]{hyperref}

\cvprfinalcopy 

\def\cvprPaperID{9801} 

\ifcvprfinal\fi
\begin{document}

\title{Can Deep Learning Recognize Subtle Human Activities?}



\author{Vincent Jacquot\\
\'{E}cole Polytechnique \\
F\'{e}d\'{e}rale de Lausanne\\
{\tt\small jacquot.vinc@gmail.com}
\and
Zhuofan Ying\\
University of Science and \\
Technology of China\\
{\tt\small zhuofanying@gmail.com}
\and
Gabriel Kreiman\\
Center for Brains, Minds \\
and Machines, Boston, MA\\
{\tt\small gabriel.kreiman@tch.harvard.edu}
}

\maketitle

\begin{abstract}
   Deep Learning has driven recent and exciting progress in computer vision, instilling the belief that these algorithms could solve any visual task. Yet, datasets commonly used to train and test computer vision algorithms have pervasive confounding factors. Such biases make it difficult to truly estimate the performance of those algorithms and how well computer vision models can extrapolate outside the distribution in which they were trained. In this work, we propose a new action classification challenge that is performed well by humans, but poorly by state-of-the-art Deep Learning models. As a proof-of-principle, we consider three exemplary tasks: drinking, reading, and sitting. The best accuracies reached using state-of-the-art computer vision models were 61.7\%, 62.8\%, and 76.8\%, respectively, while human participants scored above 90\% accuracy on the three tasks. We propose a rigorous method to reduce confounds when creating datasets, and when comparing human versus computer vision performance. Source code and datasets are publicly available\footnote{\url{https://github.com/kreimanlab/DeepLearning-vs-HighLevelVision}}.
\end{abstract}

\section{Introduction}

Deep convolutional neural networks have radically accelerated progress in visual object recognition, with impressive performance on datasets such as ImageNet \cite{ILSVRC15}, achieving top-5 error of 16.4 \% in 2012 \cite{alexnet}, down to  1.8\% in 2019 \cite{xie2019selftraining}. Similar progress has been observed in other domains such as action recognition, with an error rate of 1.8\% \cite{carreira2017quo} in the UCF101 dataset \cite{soomro2012ucf101}. 

\begin{figure}[ht]
\begin{center}
   \includegraphics[width=0.95\linewidth]{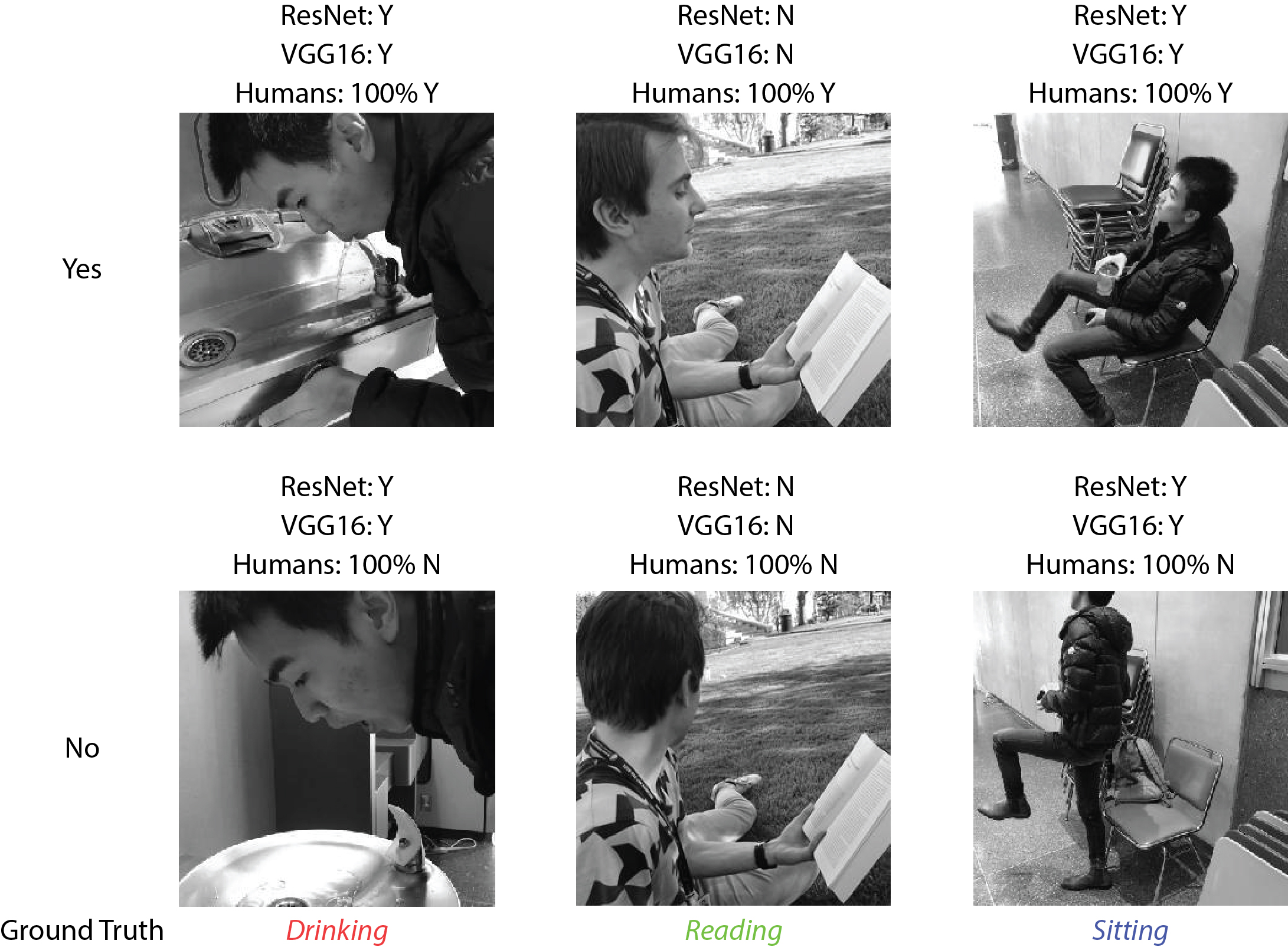}
\end{center}
   \caption{\textbf{Example images from our dataset (Group 2, controlled set)}. Left to right: \textit{drinking}, \textit{reading}, and \textit{sitting}. Top: positive images. Bottom: negative images. Above each image, classification output for ResNet, VGG16, and human psychophysics measurements (see text for details). The models misclassified the middle top, bottom left, and bottom right pictures, whereas humans correctly classified all the pictures. See also \textbf{Fig. S4}.}
\label{fig:example_imgs}
\end{figure}

Such impressive feats have also been accompanied by vigorous discussions to better understand what the networks learn and how they classify the images \cite{zeiler2013visualizing,lin2015visualizing,samek2017explainable,Montavon_2018,kindermans2017learning}. In addition to showcasing algorithmic successes, systematically understanding the networks' limitations will help us develop better and more stringent datasets to stress test models and develop better ones. For example, in the UCF101 dataset, algorithms can rely exclusively on the background color to classify human activities well above chance levels. For example, ``sky diving'' typically correlates with blue pixels (the sky), whereas `baseball pitch'' correlates with green pixels (the field).

As an illustration of how to rigorously test state-of-the-art models, and how to build controlled datasets, we focus on action recognition from individual frames. We study three human behaviors: whether a person is \emph{drinking} or not, \emph{reading} or not, and \emph{sitting} or not (\textbf{Figure \ref{fig:example_imgs}}, \textbf{Fig. S4}). Each of these actions is considered independently in a binary classification task. We first describe how we built a controlled dataset, next we demonstrate that humans can  rapidly solve these tasks, and finally we show that these simple binary questions challenge current systems, and introduce initial thoughts on how such tasks could be solved.

\section{Related Work} \label{section:rel_work}

\textbf{Object detection.} Large datasets for object detection  have played a critical role in recent progress in computer vision. 
The success of Krizhevsky \etal \cite{alexnet} on ImageNet \cite{ILSVRC15} triggered the development of powerful algorithms \cite{xie2019selftraining, touvron2019fixing, Mahajan_2018_ECCV}, and multiple datasets such as COCO \cite{lin2014microsoft}. 

\textbf{Action recognition.}  In a similar fashion, multiple datasets have been developed to train algorithms to recognize actions, including the MPII Human Pose  \cite{andriluka14cvpr}, COCO keypoints \cite{lin2014microsoft}, Leeds Sports Pose \cite{Johnson10}, UCF101 action \cite{soomro2012ucf101}, and Posetrack \cite{PoseTrack} datasets. These datasets led to the current state-of-the art models for human pose estimation \cite{Toshev_2014, tompson2014efficient, wei2016convolutional, carreira2015human, alej2016stacked, cao2018openpose}.

\textbf{Current challenges and possible approaches.} 
There has been significant progress in developing enhanced algorithms for recognition combining region proposal \cite{girshick2013rich,girshick2015fast,dai2016rfcn,he2017mask,Detectron2018}, distinction between foreground/background and other scene elements \cite{lin2017focal,ronchi2015describing, kendall2015bayesian, Detectron2018}, and interactions between image parts \cite{gkioxari2017detecting}. 

Despite enormous progress triggered by these datasets, there exist strong low-level biases that correlate with the labels. For example, the work of Xiao \etal showed that a simple architecture, combining ResNet with several deconvolution layers, reached the top accuracy of 73.7\% mAP in human pose estimation and tracking \cite{xiao2018simple}. This type of challenge is particularly notable in datasets like UCF-101: extracting \emph{merely the first frame} of each video, converting it to grayscale, and using an SVM classifier with a linear kernel, it is possible to obtain performance levels well above chance in ``action recognition''. To capitalize on the power of current algorithms, and to push the development of even better ones, it is essential to stress test computer vision systems with sufficiently well-controlled datasets that cannot be solved by simple heuristics. Here we focus on the problem of action recognition from static images and provide intuitions about the development of a well-controlled dataset to challenge computational algorithms.


\section{Building a Controlled Dataset}

We sought to create a dataset to challenge and improve current recognition algorithms, focusing on action recognition from single frames in three examples: drinking, reading, and sitting. Datasets that involve discriminating among completely different actions (as in UCF-101, \cite{soomro2012ucf101}), often incorporate extensive background information that can help solve the discrimination problem by capitalizing on basic image heuristics (as noted in the introduction for the example of skydiving versus baseball pitch). Therefore, here we take a different approach and focus on binary tasks of the form: is the person drinking or not, reading or not, sitting or not. We do \emph{not} compare drinking to reading to sitting (i.e., vertical and not horizontal comparisons in  \textbf{Figure \ref{fig:example_imgs}}).

\subsection{Dataset collection} \label{section:img_source}

The images originated from two sources: (Group 1) Photographs manually downloaded from open source materials on the Internet; (Group 2) New custom photographs taken by investigators in our lab. 

Despite our best efforts, we quickly realized that Group 1 (internet images) contained strong biases: even an SVM with a linear kernel applied to the image pixels could classify images with higher-than-chance accuracy. Consequently, we decided to take our own photographs (Group 2, controlled set, \textbf{Figure \ref{fig:example_imgs}}, \textbf{Fig. S4}). Special care was taken to avoid biases when taking pictures. Whenever we took a photo representing a behavior in a certain setting (e.g., person A drinking from a cup in location L), we also took a companion photo of the opposite behavior in the same setting (person A holding the same cup in location L but \emph{not} drinking). Examples of these image pairs for each behavior are shown in \textbf{Figure \ref{fig:example_imgs}}. The opposite behavior could be a slight change, for example the same picture with and without water in the case of \textit{drinking}, or changing the direction of gaze for \textit{reading}, or changing body posture for \textit{sitting}. This procedure ensured that the differences between the two classes could not  be readily ascribed to low-level properties associated with the two labels. We reasoned that these differences between the \textit{yes} and \textit{no} classes would make the classification task difficult for current algorithms, while still being solvable by humans. We conjectured that these subtle, but critical, differences, highlight the key ingredients of what it means for an algorithm to be able to truly recognize an action.

The original number of images in the \textit{drinking}, \textit{reading} and \textit{sitting} datasets were 4,121, 3,071 and 3,684, respectively. These datasets were then split into \textit{yes} and \textit{no} classes according to the labelling procedure described in Section \ref{section:labelling}. About 85\% of each dataset consisted of our own photographs (Group 2), while the rest was from the Internet (Group 1). All images were converted to grayscale and resized to 256-by-256 pixels (except in \textbf{Fig. S1} and \textbf{Fig. S2} which show results for RGB images).

\subsection{Labelling images} \label{section:labelling}

We created ground truth labels for each image by asking 3 participants to assign each image to a \textit{yes} or \textit{no} class for each action. The participants were given simple guidelines to  define each action: \textit{drinking} (liquid in mouth), \textit{reading} (gaze towards text), and \textit{sitting} (buttocks on support). In contrast to the psychophysics tests in Section \ref{section:psiturk}, here the 3 participants had no time constraint to provide labels. We only kept an image if all the participants agreed on the class label.

\begin{figure*}[!ht]
\begin{center}
\includegraphics[width=0.9\linewidth]{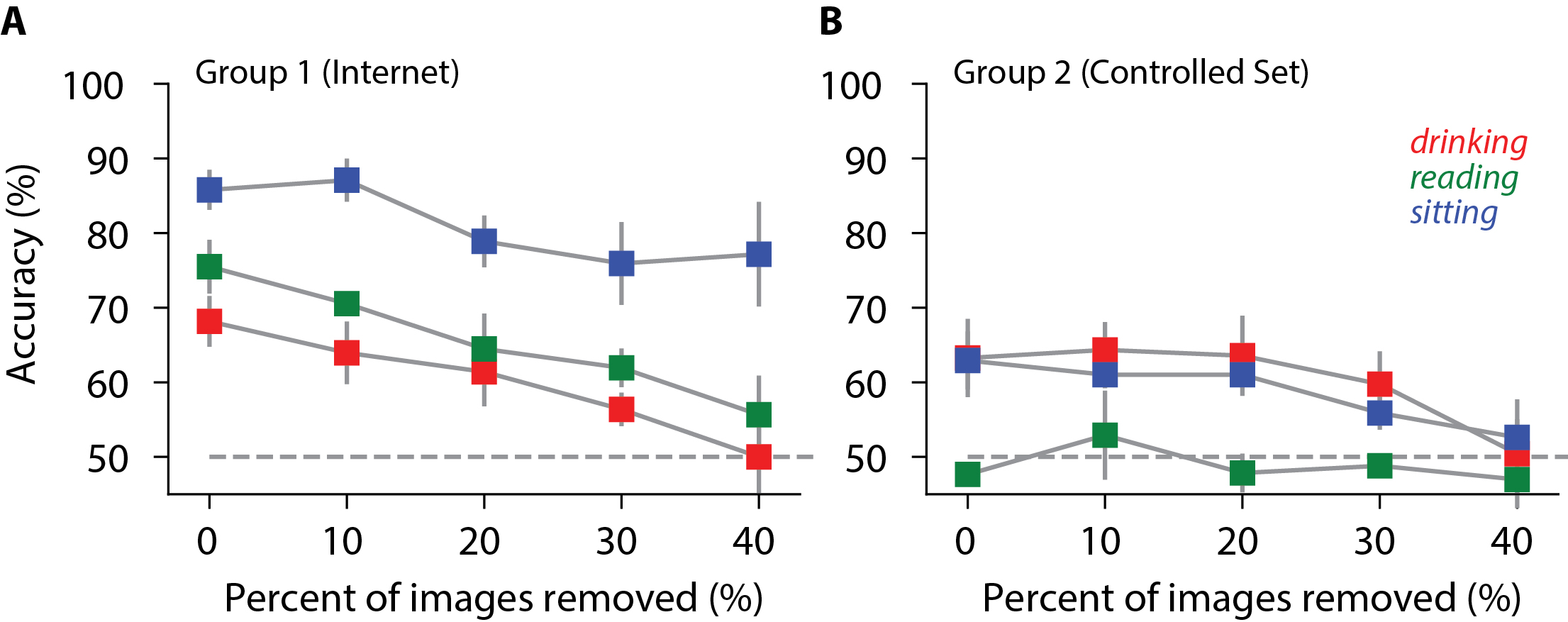}
\end{center}
   \caption{\textbf{Images downloaded from the internet carry large biases}. 
   Accuracy on the three datasets (red=\textit{drinking}, green=\textit{reading}, blue=\textit{sitting}) as a function of the percentage of images removed for images from Group 1 (\textbf{A}, Internet) or Group 2 (\textbf{B}, Controlled Set). Accuracy refers to classification results on test data using an SVM classifier on the fc7 activations of a fine-tuned AlexNet ( Section \ref{section:rmvEasyImgs}). Error bars = standard deviation. Horizontal dashed line = chance level.}
\label{fig:rmvEasyImgs}
\end{figure*}

\subsection{Removing biases} \label{section:rmvEasyImgs}

As noted in the Introduction, spurious correlations between images and labels can render tasks easy to solve. To systematically avoid such biases, we implemented a pruning procedure by ensuring that the images could not be easily classified by "simple" deep learning algorithms. This was done by applying 100 cross-validation iterations (80\%/20\%) of a fine-tuned AlexNet \cite{alexnet, malabAlexNet} on each dataset. The weights were pre-trained on ImageNet \cite{malabAlexNet}. A 2-unit fully-connected layer was added on top of the fc7 layer. Classification was performed by a softmax function using cross-entropy for the cost function. Weights were updated over 3 epochs, via Stochastic Gradient Descent (SGD) with momentum 0.9, L2 regularization with $\lambda = 10^{-4}$, and learning rate $10^{-4}$. 

After fine-tuning, an SVM was applied to the fc7 layer of fine-tuned AlexNet activations to classify the images.  Images were ranked from easiest (correctly classified in most of the 100 iterations) to hardest (correctly classified only in 50\% of the iterations). We progressively removed images from the dataset according to their rank and re-applied the same procedure on the reduced datasets. \textbf{Figure \ref{fig:rmvEasyImgs}} shows the resulting drop in accuracy, as a function of the percentage of images removed. 

Images from Group 1 (Internet) were easily classified (Figure  \ref{fig:rmvEasyImgs}A): accuracy was $68.2\pm 3.4$\%  (\textit{drinking}), $75.7 \pm 3.6$\% (\textit{reading}), and $85.8 \pm 2.7$\%  \textit{sitting}), where chance is 50\%, consistent with the biases inherent to Internet images. For example, the \textit{drinking} dataset contained images of babies in the positive but not in the negative class. Other biases could be due to the surrounding environment: positive examples of \textit{sitting} tended to correlate with indoor pictures, whereas negative examples tended to be outdoors. After eliminating 40\% of the images, \textit{drinking} reached an accuracy of $50 \pm 5.0$\%, and \textit{reading} reached an accuracy of $55.7 \pm 5.2$\%. In the case of \textit{sitting}, we had to remove up to 70\% of images to obtain close to chance-level accuracy.

The Group 2 dataset (our own photographs) was more difficult to classify (Figure \ref{fig:rmvEasyImgs}B), even without any image removed: accuracy was $63.3 \pm 5.2$\% (\textit{drinking}), $47.7 \pm 0.8$\% (\textit{reading}), and $62.9 \pm 3.9$\% (\textit{sitting}). After eliminating 40\% of the images, \textit{drinking} reached an accuracy of $50.4 \pm 7.3$\%, and \textit{sitting} reached an accuracy of and $52.6 \pm 2.4$\%, while \textit{reading} dataset remained close to chance ($50$\%).

\subsection{Final dataset}
After the processes in Sections \ref{section:labelling} and \ref{section:rmvEasyImgs}, we obtained a final dataset for each action: 2,164 images for \textit{drinking}, 2,524 images for \textit{reading}, and 2,116 images for \textit{sitting}, with 50\% \emph{yes} labels. These quantities are of the same order of magnitude as the number of images per category in the popular ImageNet dataset, where every class contains between 450 and slightly over 1,000 images. ImageNet contains many more classes (1,000 instead of the 3 x 2 classes used here). However, we note that the goal in most analyses of ImageNet is to discriminate \emph{between} different classes. Here we are interested in detecting each action in a binary yes/no fashion, and we are not trying to discriminate one activity (e.g., \textit{drinking}) from the others (e.g., \textit{sitting} or \textit{reading}). Each dataset is split into a \textit{training} set ($80$\%), \textit{validation} ($10$\%), and \textit{test} set ($ 10$\%). The persons appearing in the photographs of each set are uniquely present in that set. For example, if one person is in the \textit{training} set, then they are not present in either the \textit{validation} or \textit{test} sets.


\begin{figure}[ht]
\begin{center}
   \includegraphics[width=1\linewidth]{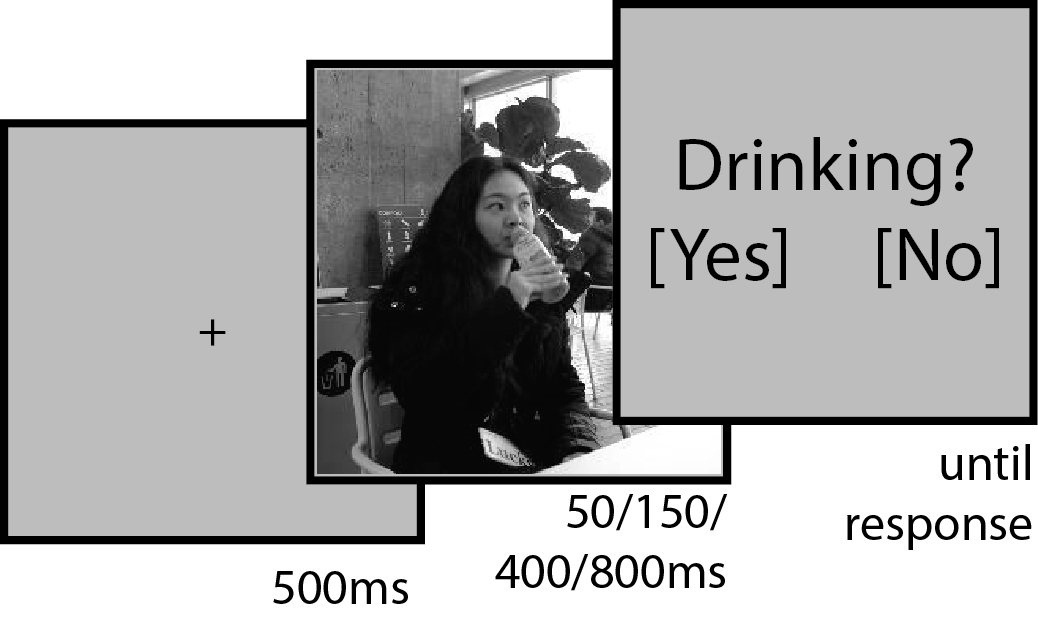}
\end{center}
   \caption{\textbf{Schematic description of the psychophysics task}  (Section \ref{section:psiturk}). Gif files were presented to mturk workers; each trial consisted of fixation (500 ms), image presentation (50, 150, 400, or 800 ms), and a forced choice yes/no question. }
\label{fig:ex_psiturk}
\end{figure}

\section{Psychophysics evaluation} \label{section:psiturk}

Ground truth labels were obtained based on the consensus of three subjects who examined the images with no time limit (Section \ref{section:labelling}). To compare human versus machine performance, we conducted a separate psychophysics test with limited exposure duration of 50, 150, 400, or 800 ms in a two-alternative forced choice task implemented with psiTurk \cite{psiTurk} (\textbf{Figure ~\ref{fig:ex_psiturk}}). The test was delivered to a total of 54 subjects via Amazon Mechanical Turk. 

\begin{figure*}
\begin{center}
\includegraphics[width=0.8\linewidth]{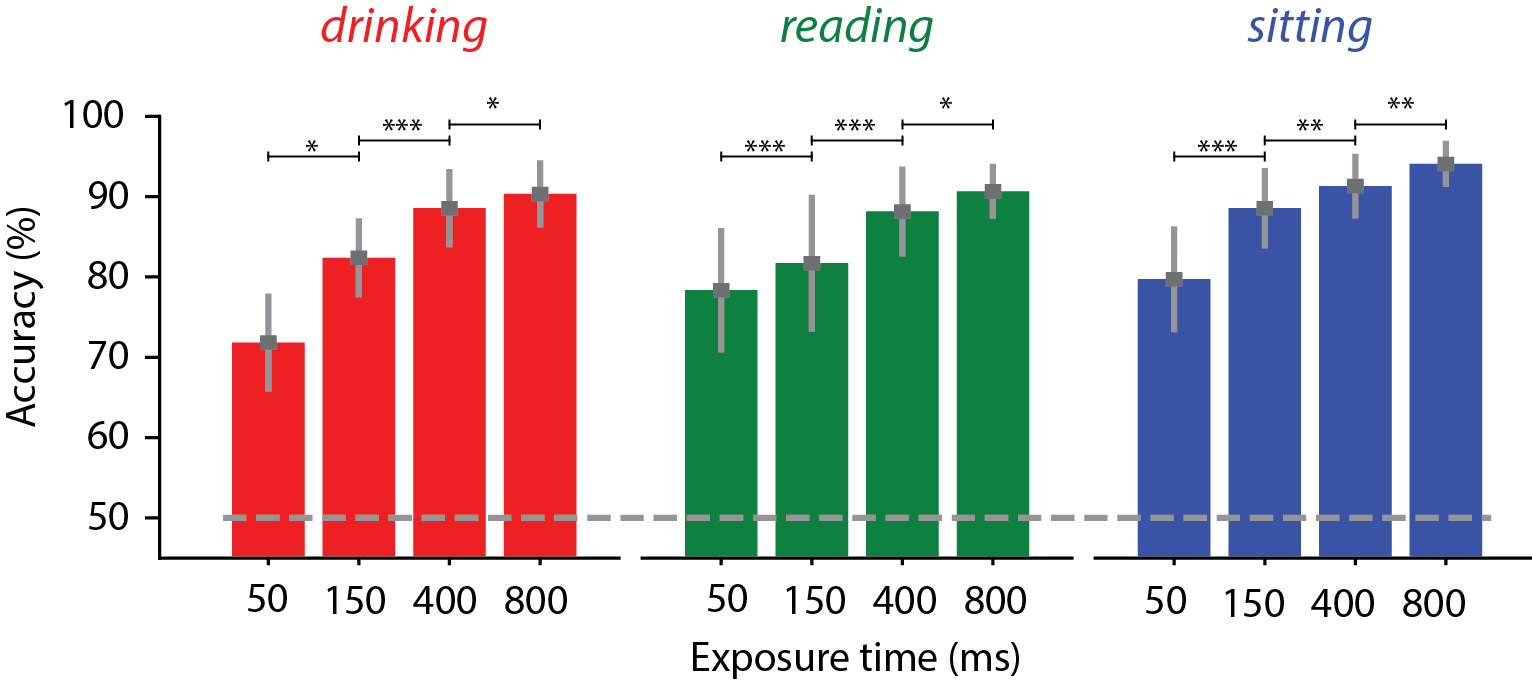}
\end{center}
   \caption{\textbf{Humans can rapidly detect the three actions}. Average accuracy $\pm$ SD as a function of exposure time on the three datasets in the task shown in \textbf{Figure \ref{fig:ex_psiturk}}. (***) $p < 0.0005$, (**) $p < 0.05$, (*) $p < 0.1$ on one-tailed, paired t-test. Horizontal dashed line = chance level.}
\label{fig:psiturk_results}
\end{figure*}

The trial sequence was presented as .gif files to approximately control the duration of image presentation (Figure \ref{fig:ex_psiturk}). Each trial consisted in a fixation cross (500 ms), followed by the image presented for a duration of either 50, 150, 400 or 800 ms, and finally a two-alternative forced choice question shown until the subject answered \cite{tangetal2018}. The image duration changed randomly from one presentation to the next.
Despite selecting only ``master mturk workers'' with a rate of past accepted hits higher than 99\%, online experiments often have subjects who do not fully attend or understand the task. To avoid including such cases, outlier subjects that showed a significantly lower accuracy than the population (p-value $<$ 0.05 on one-tailed $t$-test) were excluded from further analyses. This threshold concerned  3 out of 18 (\textit{drinking}), 3 out of 19 (\textit{reading}), and 2 out of 17 (\textit{sitting}) subjects. 

The average accuracy as a function of image duration for the human subjects is shown in \textbf{Figure \ref{fig:psiturk_results}}. Even at the shortest duration (50 ms), subjects were significantly above chance in all tasks, with a performance of at least $71.8\pm6.1$\% (\textit{drinking}), up to $79.7 \pm 6.6$\% (\textit{sitting}). As expected, performance increased with exposure time. At the longest duration of 800 ms, performance was above $90$\% for all three tasks. 

\section{State-of-the-art models}

We considered two main families of strategies to solve the task: (1) We used state-of-the-art deep convolutional neural networks pre-trained on the ImageNet dataset \cite{ILSVRC15}, with or without fine-tuning on the current dataset (\ref{section:imagenet-models}); and (2) extraction of putative action-relevant features using the Detectron algorithm \cite{Detectron2018}, a state-of-the-art object-detection algorithm pre-trained on the COCO dataset \cite{lin2014microsoft}.

\begin{figure*}
\begin{center}
\includegraphics[width=1\linewidth]{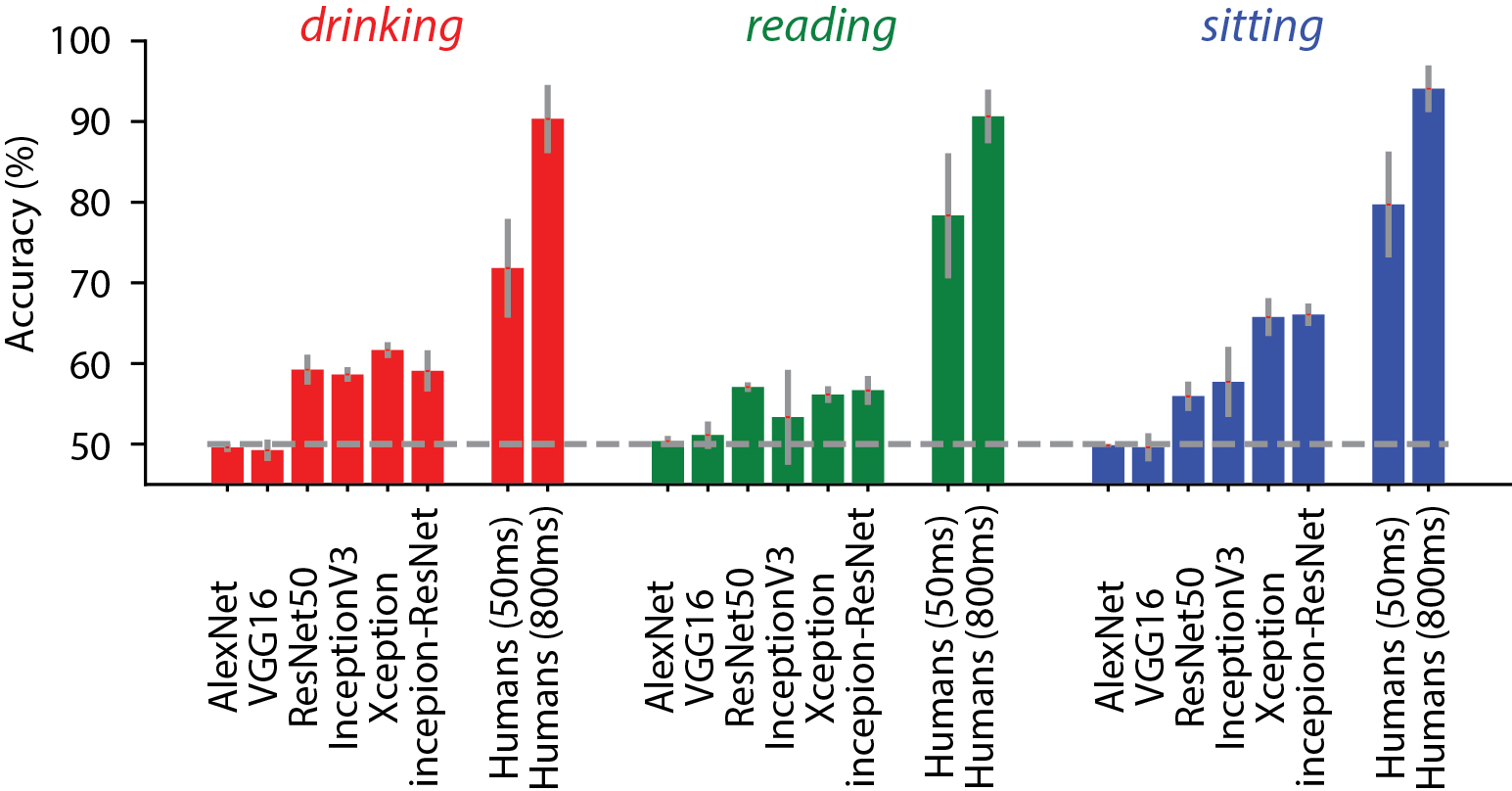}
\end{center}
   \caption{\textbf{Deep convolutional neural network models were far from human-level performance}.
   Test performance for each fine-tuned model is shown ($mean \pm SD$). The model with best accuracy on the validation set was retained to be applied on the test set, as described in section \ref{section:imagenet-models}.
   We also reproduce here the human performance values for 50ms and 800ms exposure from \textbf{Figure \ref{fig:psiturk_results}} for comparison purposes.
   Human accuracy was significantly better than any of the algorithms, ($p< 0.0005$, one-tailed $t$-test). Horizontal dashed line = chance performance.}
\label{fig:final_results_all}
\end{figure*}

\subsection{Models pre-trained on ImageNet and fine-tuned on the current dataset} \label{section:imagenet-models}

We considered the following deep convolutional neural networks: AlexNet \cite{alexnet}, VGG16 \cite{simonyan2014deep}, InceptionV3 \cite{DBLP:journals/corr/SzegedyVISW15}, ResNetV2 \cite{DBLP:journals/corr/HeZR016}, Inception-ResNet \cite{szegedy2016inceptionv4} and Xception \cite{DBLP:journals/corr/Chollet16a} available from Keras \cite{chollet2015keras}. Weights were pre-trained on ImageNet. The last classification layer, made of 1,000 units for ImageNet, was replaced by a 512x1 fc layer, followed by a 1-unit classification layer. All weights were updated via Adam optimization \cite{kingma2014adam}, with a learning rate of $10^{-4}$, until validation accuracy stagnated. Cost was measured with binary cross-entropy and the classifier was Softmax. 

We first considered the pre-trained weights followed by a classification layer. We next considered  fine-tuning only the last layers. We finally considered fine-tuning the entire network with the images in the current dataset. The model yielding the highest accuracy on the validation set was applied to the test set. Results are shown in Figure \ref{fig:final_results_all}. The top accuracy on the \textit{drinking} dataset was $61.7 \pm 0.9$\%, obtained with the Xception network \cite{DBLP:journals/corr/Chollet16a}. This is far below the 90.3\% accuracy reached by humans on this task. Inception-ResNet \cite{szegedy2016inceptionv4} gave the best results for \textit{reading} and \textit{sitting}, with $56.7 \pm 1.8$\% and $66.1 \pm 1.4$\% accuracy respectively. These values are also far below the 90.7\% and 94.1\%, respectively, reached by humans.

We tested several additional variations in an attempt to improve performance. First, using RGB images instead of grayscale images led to similar performance, well below the accuracy obtained by humans using grayscale images (\textbf{Figure S1}). In contrast to uncontrolled datasets where color can provide strong cues (as in the skydiving versus baseball pitch example noted in the Introduction), in a more controlled dataset color does not help much. Second, accuracy was slightly improved using artificial data augmentation. Every image was horizontally flipped with probability 50\%, and shifted along x or y axis by a number of pixels randomly picked in the interval [-30,30] \cite{chollet2015keras}. Third, several regularization techniques were evaluated but neither L1 nor L2 normalization improved the accuracy. Finally, replacing the penultimate 512-unit fully-connected layer by 1,024 units with drop-out did not improve the accuracy either. In sum, none of the networks and variations tested here were close to human performance, even when forcing humans to use grayscale images and respond after 50 ms exposure. 

We visualized the salient features relevant for classification in these networks using Grad-CAM \cite{SelvarajuDVCPB16}. \textbf{Figure S3} shows an example visualization for the ResNet-50 network \cite{DBLP:journals/corr/HeZR016} with weights pre-trained on ImageNet. Even though the networks often (but not always) focused on relevant parts of the image (such as the mouth or hands for drinking), the models failed to capture the critical nuances in each image that distinguish each action. For example, reading critically depends on assessing whether the gaze is directed towards text or not.

\subsection{Extraction of putative action-relevant features} \label{section:detectron}

Despite using a variety of state-of-the-art deep convolutional neural network architectures, with or without fine-tuning, colors, different regularizers, or data augmentation, humans outperformed all the algorithms by a large amount (\textbf{Figure \ref{fig:final_results_all}}). 

We reasoned that humans may capitalize on additional knowledge about the specific elements and interactions between elements that are involved in defining a given action. For example, \textit{reading} depends on the presence of text (a book, a magazine, a sign), a person, and gaze directed from the person toward the text. To test this idea, we applied algorithms where we could impose the definition of each action by using computational approaches to detect the corresponding elements and their interactions. 

\begin{figure*}
\begin{center}
\includegraphics[width=0.8\linewidth]{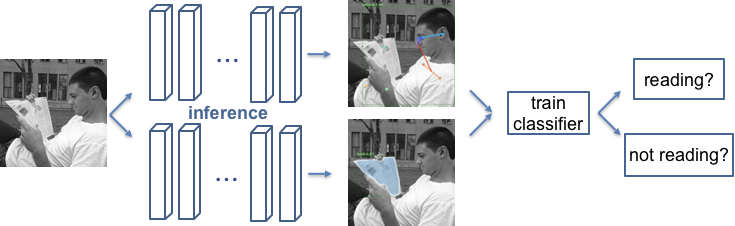}
\end{center}
   \caption{\textbf{Action-dependent extraction of relevant keypoints and features for \textit{reading}.}
   Schematic of the implementation of Detectron \cite{Detectron2018}, as described in Section \ref{section:detectron}. On the reading dataset, we combined two implementations of Detectron. Top: Detectron trained on the Keypoint dataset of COCO \cite{lin2014microsoft} allows to extract features, keypoints and bounding-box of the person in the image. Bottom: Detectron trained on the Object Detection dataset of COCO allows to extract the bounding-box and features of the reading material in the picture (see text for details). }
\label{fig:detectron_im}
\end{figure*}

We employed  two implementations of the Detectron algorithm \cite{Detectron2018} to pursue this approach (\textbf{Figure \ref{fig:detectron_im}}). In the first approach (Model A), we used the Detectron X-101-32x8d-FPN\_s1x configuration, where 32x8d means 32  groups  per  convolutional layer and a bottleneck width of 8 \cite{xie2016aggregated}, while s1x refers to the slow learning-rate schedule. This model was trained on the Keypoint Detection Task from the COCO dataset \cite{lin2014microsoft}, comprising 150,000 person instances labelled with 17 keypoints covering their body (ankles, knees, elbows, eyes, among other points).

In the second approach (Model B), we used the Detectron X-101-64x4d-FPN\_1x configuration (64 convolutional groups with a bottleneck width of 4). This model was trained for the Object Detection Task of the COCO dataset \cite{lin2014microsoft}, consisting of 82,000 images with the objective of segmenting 81 classes of objects.

Both implementations use Mask R-CNN \cite{he2017mask} and Feature Pyramid Network \cite{lin2016feature} for the architecture, with 101-layers ResNeXt as a backbone \cite{xie2016aggregated}. Both implementations obtain the highest performance in their respective tasks.

\begin{figure*}
\begin{center}
\includegraphics[width=1\linewidth]{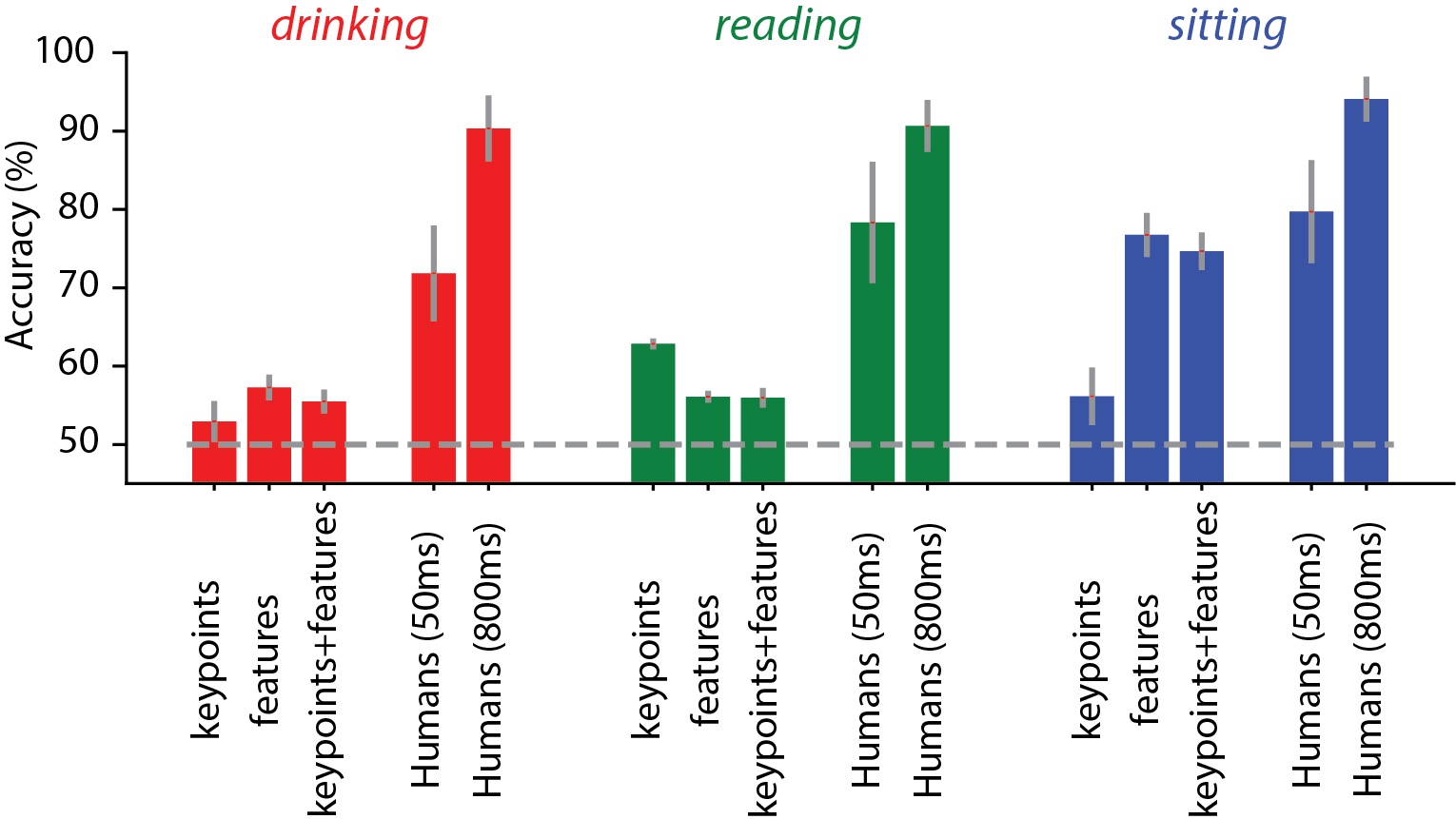}
\end{center}
   \caption{\textbf{Extracting action-relevant features can improve performance but all models remain well below human levels}. We extracted specific \emph{keypoints} and \emph{features} using the Detectron algorithm (see \textbf{Figure \ref{fig:detectron_im}}, and text for details). The combination of action-specific keypoints and relevant object features improved performance with respect to the architectures studied in \textbf{Figure~\ref{fig:final_results_all}} for the reading and sitting datasets. Human performance with 50ms and 800ms exposure is reproduced here from \textbf{Figure \ref{fig:psiturk_results}} for comparison purposes. Horizontal line = chance performance. None of these models reached human performance levels. }
\label{fig:detectron}
\end{figure*}

For \textbf{sitting}, only Model A was used. We extracted the bounding box, keypoints and the features of the “main” person in the picture. We defined the “main” person as the largest bounding-box whose probability of belonging to the class \textit{person} was higher than a threshold set in the implementation. Out of the extracted data, we created two vectors: a \textit{features} vector, made of the 12,544 features associated with the person in the picture, and a \textit{keypoints} vector. The \textit{keypoints} vector consisted of the x-coordinate, y-coordinate, the probability of each detected keypoint, plus the width and height of the \textit{person} bounding-box. This resulted in a vector of 53 elements, which were normalized with respect to the bounding box coordinates. A 3 fc-layer neural network (512x1, 512x1, 2x1), trained with stochastic gradient-descent, provided the best results from the \textit{features} vector while an SVM classifier was best for the \textit{keypoints} vectors. The best accuracy was $76.7 \pm 2.8$\%, obtained from the \textit{features} vectors. Grouping the two vectors together did not increase accuracy (\textbf{Figure \ref{fig:detectron}}).

For \textbf{reading}, we used both models A and B. Model A was used to extract the bounding box, keypoints and the features of the main person in the picture, similarly to the \textit{sitting} task. We used model B to extract the bounding box and features of the text material. We selected the region of interest whose probability of belonging to the classes tv, laptop, cell phone or book was higher than a certain threshold. If there were several such items in a picture, we retained the one with the largest bounding box. We combined the features from both models A and B into \textit{features} vectors. Keypoints from models A and B were grouped into \textit{keypoints} vectors. The same classifiers as for \textit{sitting} were used. The best performance was reached from \textit{keypoints} vectors with $62.8\% \pm 0.7$\% accuracy, \textit{features} vectors gave $56.1\% \pm 0.7$\% accuracy.

Addressing the \textbf{drinking} task followed a similar reasoning to the \textit{reading} task described previously. We used model A to extract the bounding box, keypoints and the features of the main person in the picture. We used model B to extract the bounding box and features of the beverage. We selected the region of interest whose probability of belonging to the classes bottle, glass, or cup was higher than a certain threshold. If there were several such items in a picture, we retained the one with the largest bounding box. We combined the features from both models A and B into \textit{features} vectors. Keypoints from model A and B were grouped into \textit{keypoints} vectors. The same neural network classifier as for \textit{sitting} and  \textit{reading} was used. The best performance was reached from \textit{features} vectors with $57.3\% \pm 1.6$\%  accuracy, while \textit{keypoints} vectors gave $52.9\% \pm 2.6$\% accuracy (\textbf{Figure \ref{fig:detectron}}).

As discussed in Section \ref{section:imagenet-models}, using RGB images instead of grayscale images led to similar accuracy, with all the models still falling below human performance levels (\textbf{Figure S2}).



\section{Discussion}

Can Deep Learning algorithms learn the concepts of drinking, reading, and sitting? We consider these basic activities as paradigmatic examples of daily actions that humans can recognize rapidly and seemingly effortlessly in a wide variety of different scenarios. Exciting progress in action recognition using datasets like UCF101 \cite{soomro2012ucf101} might convey the erroneous impression that it is relatively straightforward to develop algorithms that correctly detect activities like ``playing cello'', ``breastroke'', or ``soccer juggling''. However, it is important to note that algorithms can perform well above chance levels in these datasets, even simply using a linear classifier on pixel levels using just a single frame. In this work, we propose a methodology to build better controlled datasets. As a proof-of-principle, we introduce a prototype of such a dataset for the actions of \textit{drinking}, \textit{reading}, and \textit{sitting}. Using this controlled dataset, we show that the latest artificial neural networks are likely to extract some correct discriminative features as well as biased features for these behaviors and that humans outperform all of the current networks. 

One approach followed by prominent datasets like ImageNet \cite{ILSVRC15} or UCF101 \cite{soomro2012ucf101} is to collect example images from internet sources for a wide variety of different classes is. This approach is fruitful because it inherently represents to some extent the statistics of images in those internet sources, because there is some degree of variation captured in those images, because it enables studying multiple image classes, and because it is empirically practical. At the same time, this approach suffers from the biases inherent to uncontrolled experiments where many confounding variables may correlate with the variables of interest \cite{barbuetal2019}. 

Here we take a different approach whereby we consider detecting the presence or absence of specific actions. Even in this binary format, and despite our best intentions, it is difficult to download images from the internet that are devoid of biases (\textbf{Figure \ref{fig:rmvEasyImgs}A}). For example, perhaps there are more images of people reading indoors under artificial light conditions than outdoors and therefore low-level image properties can help distinguish \textit{reading} from \textit{not reading} images. These biases are not always easy to infer. Regardless of the exact nature of the biases between the two classes, it is clear that images downloaded from the Internet display multiple confounding factors. In an attempt to ameliorate such biases, we took our own set of photographs under approximately standardized conditions (\textbf{Figure \ref{fig:example_imgs}}, \textbf{Fig. S4}). This approach led to a substantial reduction in the amount of bias in the dataset (\textbf{Figure \ref{fig:rmvEasyImgs}B}), but it was not completely bias free. Therefore, we instituted a procedure to remove images that were easy to classify.  

Human subjects were still able to detect the three actions in the resulting datasets (\textbf{Figure \ref{fig:psiturk_results}}), even when exposure times were as short as 50 ms. Longer exposures led to close to ceiling performance for humans.  

Computational models pre-trained on object classification datasets performed barely above chance in the three tasks (\textbf{Figure \ref{fig:rmvEasyImgs}B}), even though the same models have been successful in the original datasets they were trained on. We re-trained state-of-the-art computational models using our datasets. Even after extensive fine tuning, data augmentation, adding color and regularizers, even the best models were well below human performance (\textbf{Figure \ref{fig:final_results_all}}). These results should not be interpreted as a proof that no deep convolutional neural network model can reach human level performance in this dataset. On the contrary, we hope that this dataset will inspire development of better algorithms that can thrive when the number of biases is significantly reduced. 
An important variable in deep convolutional neural network approaches is the amount of training data. Each of our datasets contain more than 2,000 images (that is, more than 1,000 images for the \emph{yes} and \emph{no} classes in each case). The ImageNet dataset contains between 450 and slightly more than 1,000 images in each class. The UCF101 dataset contains on the order of 100 videos for each class. Thus, the number of images per class in our dataset is comparable or larger than the ones in prominent datasets in the field. 

The total number of different tasks is very different. Here we only consider three binary tasks, whereas the typical format of object classification in ImageNet involves a single task with 1,000 classes and UCF101 involves a single task with 101 classes. Because of our binary approach, the total number of different tasks is not relevant to the results shown here. We assume that the same conclusions would apply to well-controlled datasets for other actions such as soccer juggling or not, playing cello or not, and others, but this remains to be determined. Extending our dataset creation protocol from 3 tasks to 100, or 1,000, different tasks is challenging due to the manual approach involved in taking photographs. However, recent efforts have astutely taken advantage of Amazon Mechanical Turk to collect pictures \cite{barbuetal2019}, an approach that could pave the way towards creating larger, yet adequately controlled, datasets. 

In the interest of simplicity, here we focus on action recognition from static images as opposed to video. We were inspired to focus on static images because it is easy to thrive in current action recognition challenges by ignoring the video information. However, there is no doubt that temporal information from videos can provide a major boost to performance. Video material downloaded from the Internet suffers from similar biases to the ones discussed above for static images. Additional biases may be introduced in videos (for example, certain video classes may have more camera movement than others). It would be interesting to follow a similar approach to the one suggested here to build controlled video datasets.

The mechanisms by which human observers recognize these actions are poorly understood. It is also unclear how much class-specific training humans have with these actions. It is interesting to conjecture that many actions can be defined by an agent, an object, and a specific interaction between the two. \emph{Drinking} involves a person (or animal), liquid, and a mechanism by which the liquid flows into the agent's mouth. Similarly, \emph{reading} involves a person, text, and gaze directed from the person to the text. Following up on this conjecture, we provide initial steps towards defining variables of interest for action recognition using the Detectron algorithm (\textbf{Figure \ref{fig:detectron_im}}). 

When designing experiments, scientists typically devote major efforts to minimizing possible biases and confounding factors. Building less biased datasets can help challenge existing algorithms and develop better algorithms that can robustly generalize to real-world problems.  


\section*{Acknowledgements}
\noindent This work was supported by NIH R01EY026025 and by the Center for Minds, Brains and Machines, funded by NSF STC award CCF-1231216. This work was inspired by discussions with and lectures presented by Shimon Ullman. We thank all the participants who were models in our photographs. In particular, we are grateful to Pranav Misra and Rachel Wherry who took and labeled the initial pictures. 

{\small
\bibliographystyle{ieee_fullname}
\bibliography{egbib}
}

\clearpage



\renewcommand{\thesection}{S\arabic{section}}
\renewcommand{\thefigure}{S\arabic{figure}}
\renewcommand{\thetable}{S\arabic{table}}
\setcounter{figure}{0}

\renewcommand{\textfraction}{0.01}
\renewcommand{\topfraction}{0.01}
\renewcommand{\bottomfraction}{0.01}
\renewcommand{\floatpagefraction}{0.01}
\setcounter{totalnumber}{1}

\begin{figure*}
\begin{center}
\includegraphics[width=14cm]{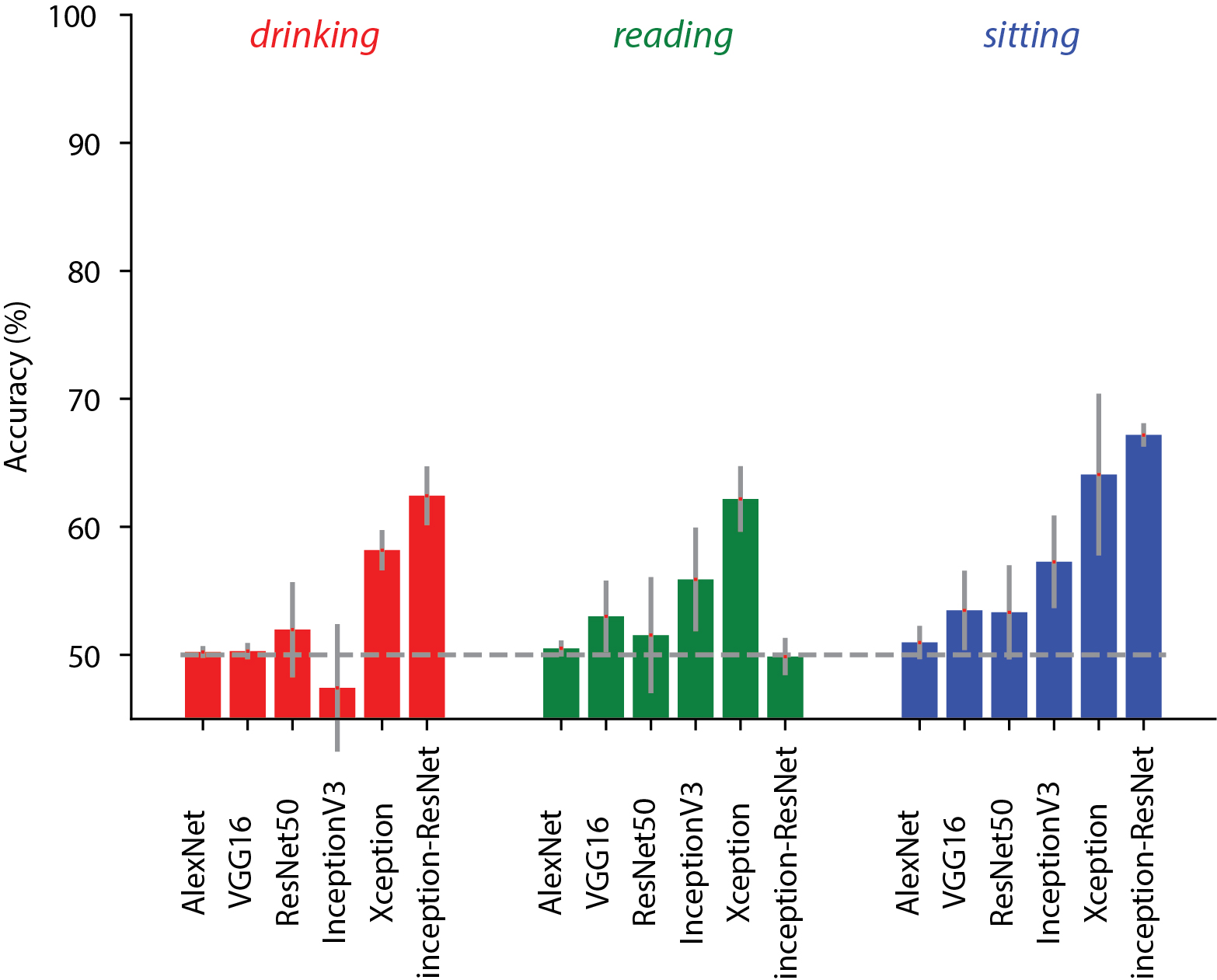}\vspace{-4mm}
\end{center}
\caption{
\textbf{Performance of deep convolutional neural network models in action recognition using RGB images}.
   This figure follows the conventions and format of \textbf{Figure 5} in the main text. Here we present results using RGB images. Test performance for each fine-tuned model is shown ($mean \pm SD$). The model with best accuracy on the validation set was retained to be applied on the test set.
 }
  \vspace{-4mm}
  \label{fig:fig_s1}
\end{figure*}

\clearpage
\newpage

\begin{figure*}
\begin{center}
\includegraphics[width=14cm]{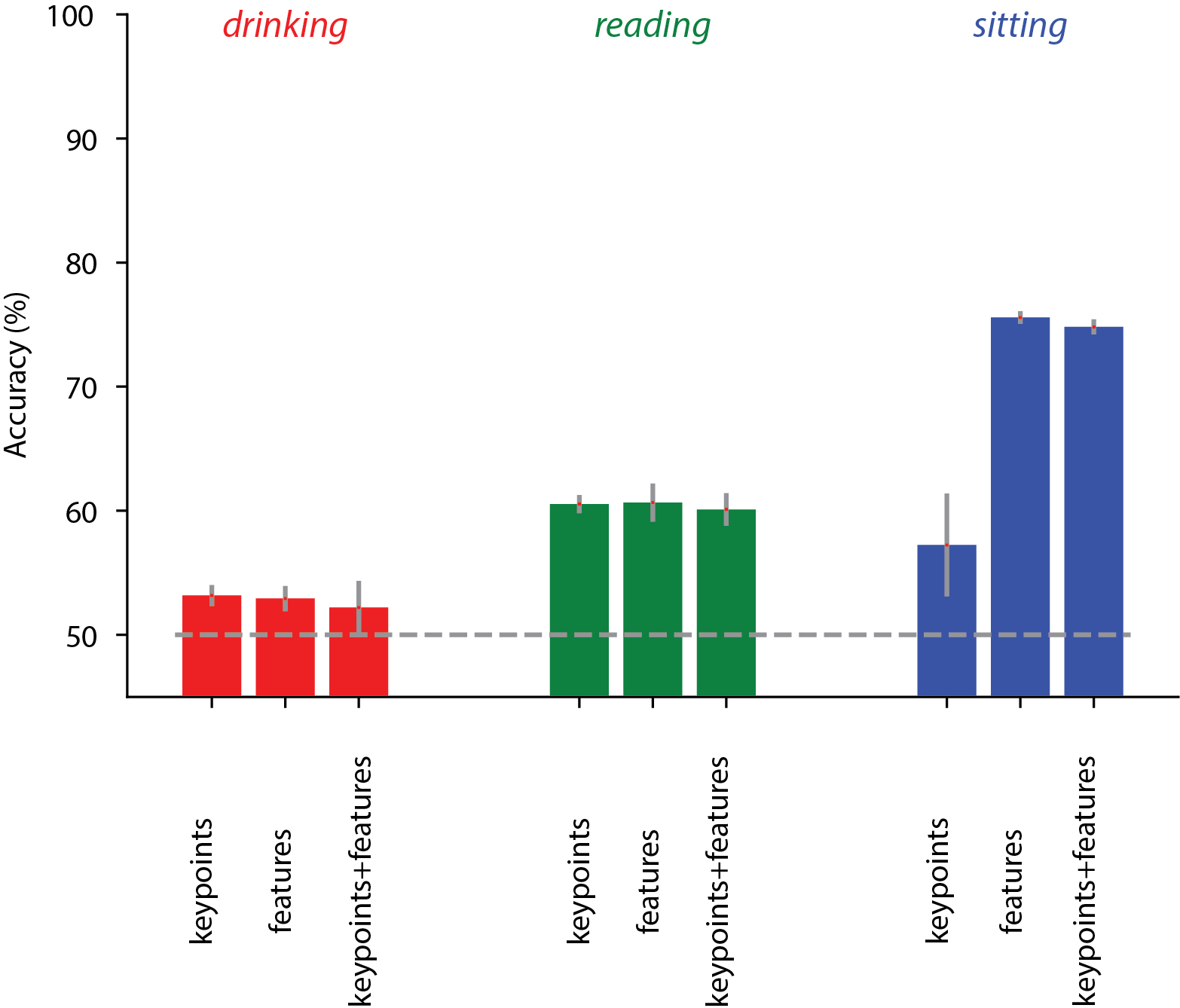}\vspace{-4mm}
\end{center}
\caption{
\textbf{Performance of detectron models extracting task-relevant features using RGB images}. This figure follows the conventions and format of \textbf{Figure 7} in the main text. Here we present results using RGB images. We extracted specific \emph{keypoints} and \emph{features} using the Detectron algorithm \cite{Detectron2018} (see main text for details). 
}
  \vspace{-4mm}
  \label{fig:fig_s2}
\end{figure*}

\clearpage
\newpage

\begin{figure*}
\begin{center}
\includegraphics[width=8cm]{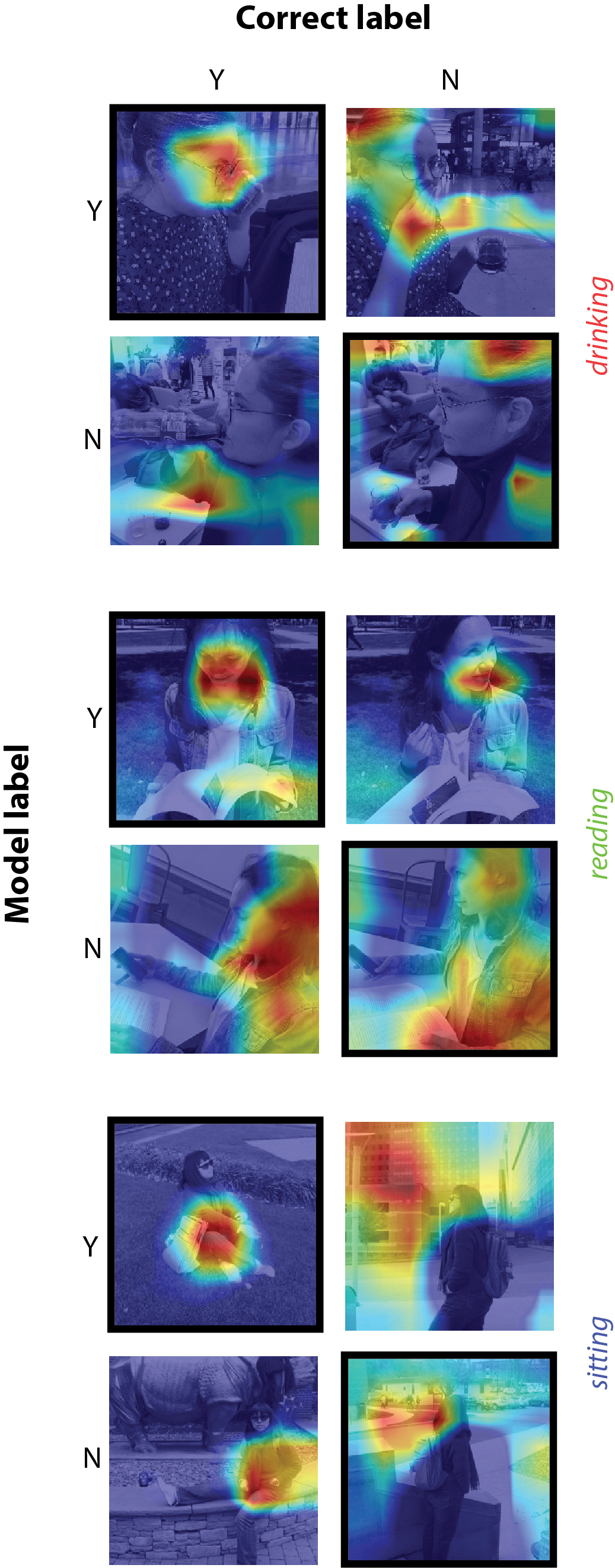}\vspace{-4mm}
\end{center}
\caption{
\textbf{Visualization of relevant features used by the network for classification}. Visualization of the salient features using Grad-CAM \cite{SelvarajuDVCPB16} for the ResNet-50 network \cite{DBLP:journals/corr/HeZR016} with weights pre-trained on ImageNet, finetuned on either the drinking, reading or sitting datasets.
The gradient is used to compute how each feature contributes to the predicted class of a picture. On the last convolutional layer, the values of the features translate to a heatmap (red for most activated, blue for least activated). The heatmap is resized from 8x8 to 256x256 such as to overlap the input image.
}
  \vspace{-4mm}
  \label{fig:visualizations}
\end{figure*}

\clearpage
\newpage

\begin{figure*}
\begin{center}
\includegraphics[width=16cm]{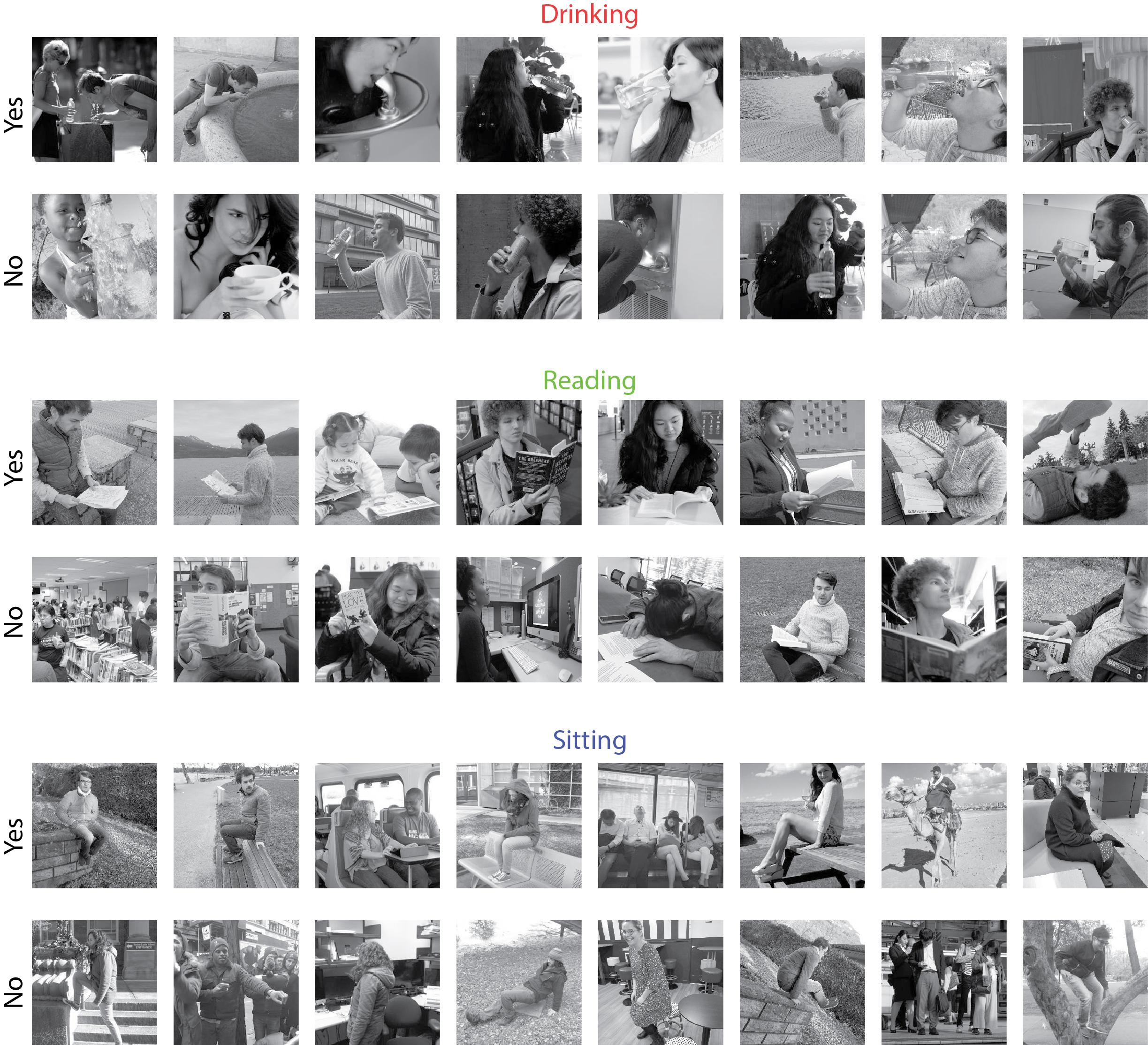}\vspace{-4mm}
\end{center}
\caption{
\textbf{Example images from our dataset.}
}
  \vspace{-4mm}
  \label{fig:visualizations}
\end{figure*}

\clearpage

\end{document}


\title{Supplementary Material \\ Can Deep Learning Recognize Subtle Human Activities?} 


\author[1]{Vincent Jacquot}
\author[2]{Zhuofan Ying}
\author[3,4]{Gabriel Kreiman}

\affil[ ]{jacquot.vinc@gmail.com,zuofanying@gmail.com,gabriel.kreiman@tch.harvard.edu}
\affil[1]{Ecole Polytechnique Federale de Lausanne}
\affil[2]{University of Science and Technology of China}
\affil[3]{Children's Hospital, Harvard Medical School}
\affil[4]{Center for Brains, Minds, and Machine}

\maketitle



\renewcommand{\thesection}{S\arabic{section}}
\renewcommand{\thefigure}{S\arabic{figure}}
\renewcommand{\thetable}{S\arabic{table}}
\setcounter{figure}{0}

\renewcommand{\textfraction}{0.01}
\renewcommand{\topfraction}{0.01}
\renewcommand{\bottomfraction}{0.01}
\renewcommand{\floatpagefraction}{0.01}
\setcounter{totalnumber}{1}

\section{Supplementary figures}

\begin{figure*}
\begin{center}
\includegraphics[width=14cm]{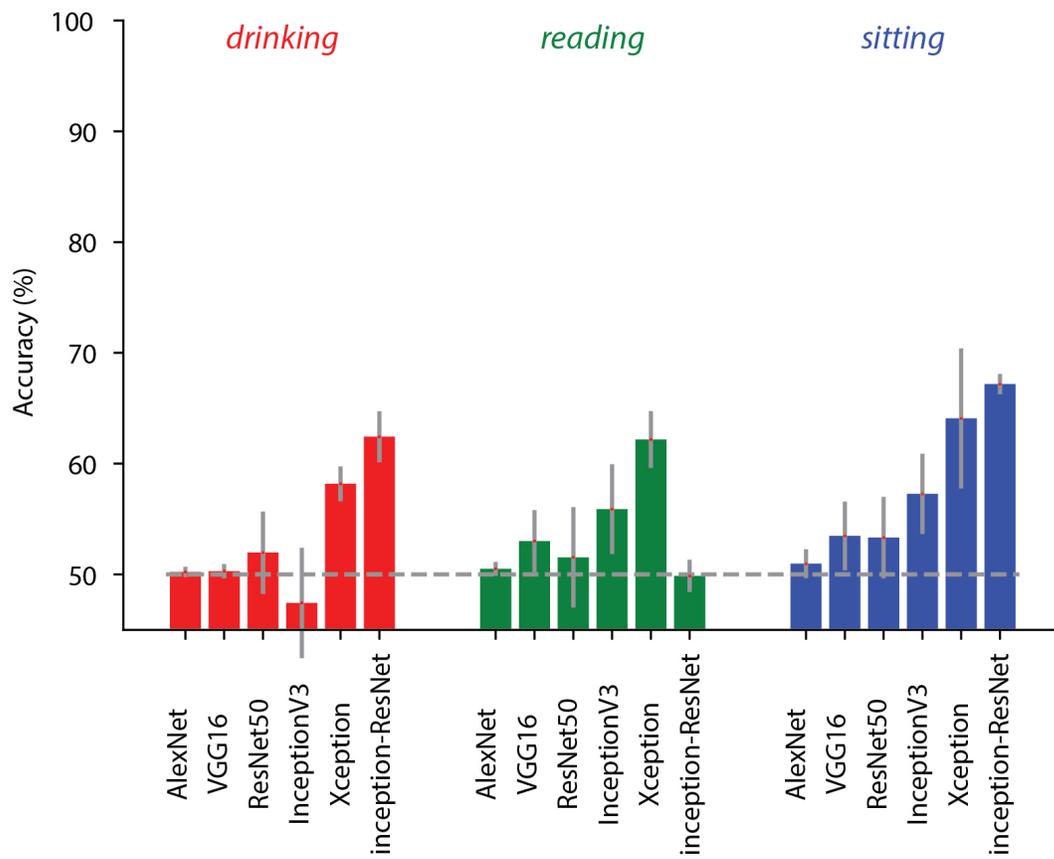}\vspace{-4mm}
\end{center}
\caption{
\textbf{Performance of deep convolutional neural network models in action recognition using RGB images}.
   This figure follows the conventions and format of \textbf{Figure 5} in the main text. Here we present results using RGB images. Test performance for each fine-tuned model is shown ($mean \pm SD$). The model with best accuracy on the validation set was retained to be applied on the test set.
 }
  \vspace{-4mm}
  \label{fig:fig_s1}
\end{figure*}

\clearpage
\newpage

\begin{figure*}
\begin{center}
\includegraphics[width=14cm]{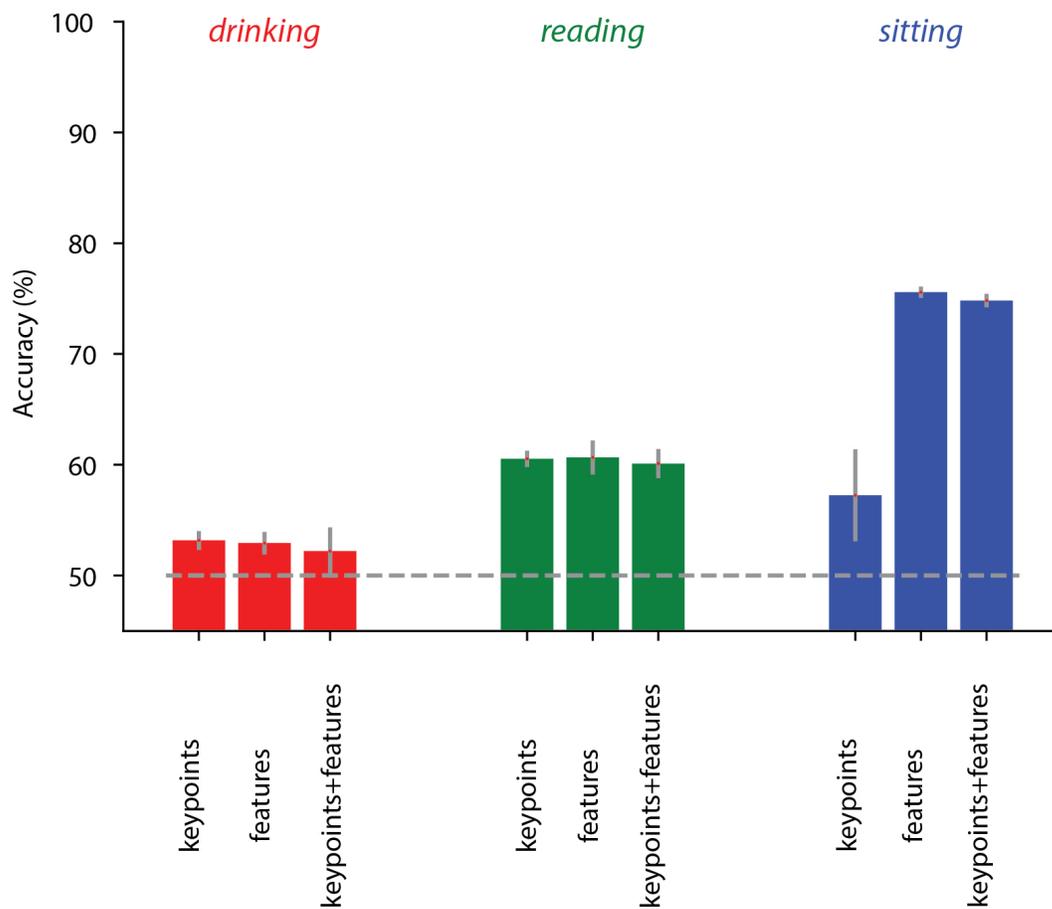}\vspace{-4mm}
\end{center}
\caption{
\textbf{Performance of detectron models extracting task-relevant features using RGB images}. This figure follows the conventions and format of \textbf{Figure 7} in the main text. Here we present results using RGB images. We extracted specific \emph{keypoints} and \emph{features} using the Detectron algorithm \cite{Detectron2018} (see main text for details). 
}
  \vspace{-4mm}
  \label{fig:fig_s2}
\end{figure*}

\clearpage
\newpage

\begin{figure*}
\begin{center}
\includegraphics[width=8cm]{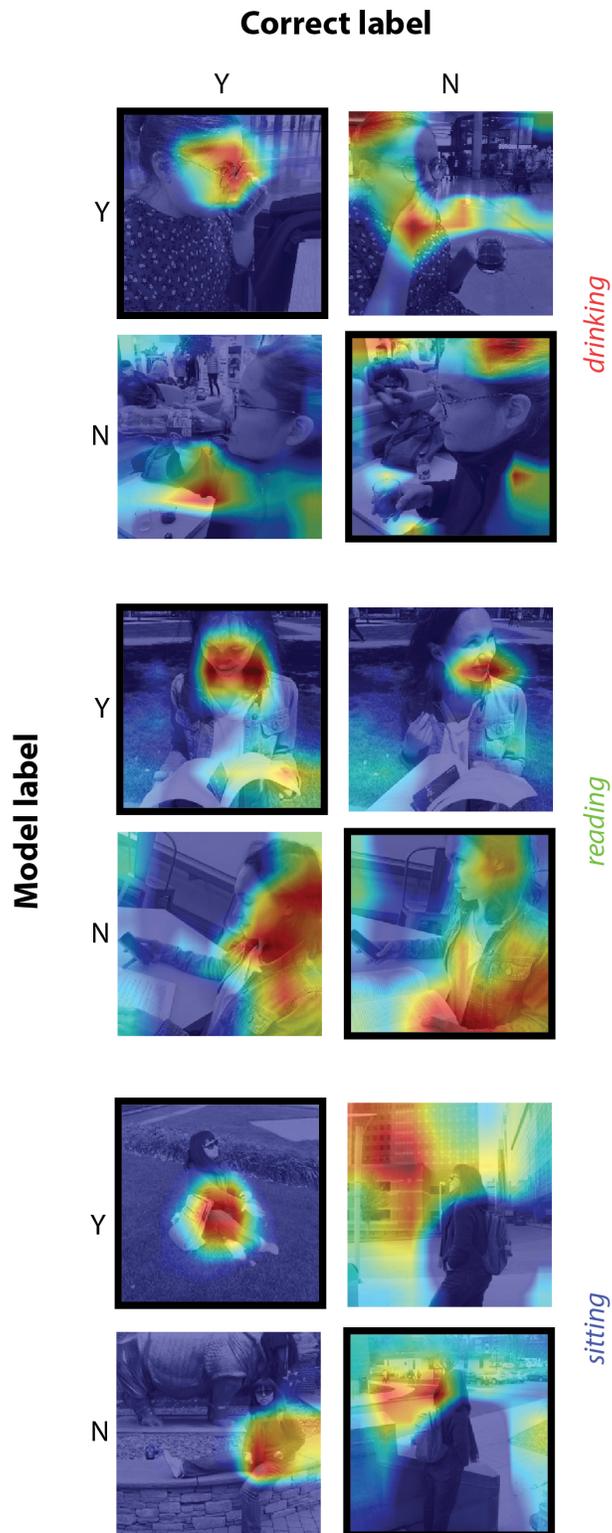}\vspace{-4mm}
\end{center}
\caption{
\textbf{Visualization of relevant features used by the network for classification}. Visualization of the salient features using Grad-CAM \cite{SelvarajuDVCPB16} for the ResNet-50 network \cite{DBLP:journals/corr/HeZR016} with weights pre-trained on ImageNet, finetuned on either the drinking, reading or sitting datasets.
The gradient is used to compute how each feature contributes to the predicted class of a picture. On the last convolutional layer, the values of the features translate to a heatmap (red for most activated, blue for least activated). The heatmap is resized from 8x8 to 256x256 such as to overlap the input image.
}
  \vspace{-4mm}
  \label{fig:visualizations}
\end{figure*}

\clearpage
\newpage

\begin{figure*}
\begin{center}
\includegraphics[width=16cm]{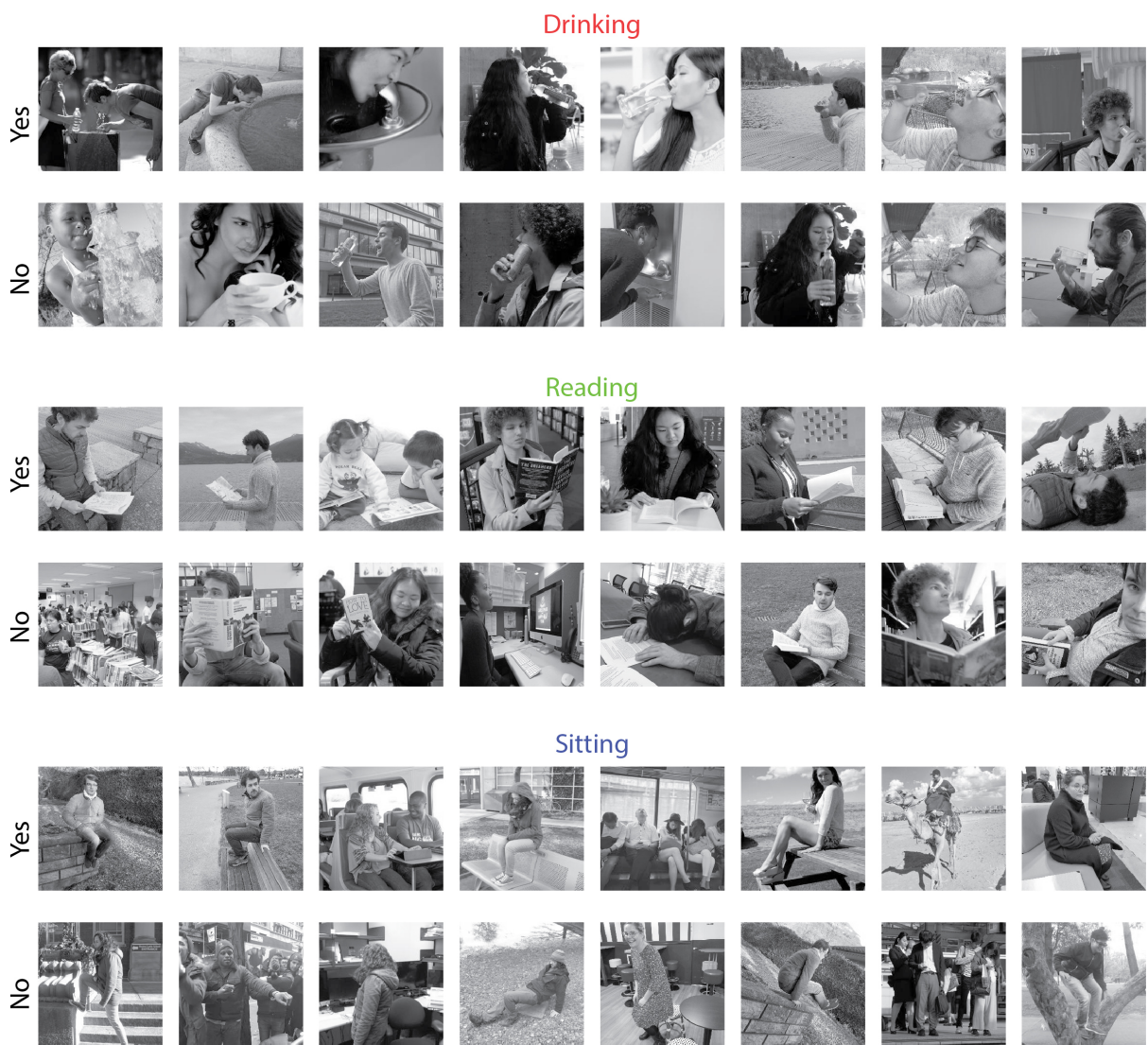}\vspace{-4mm}
\end{center}
\caption{
\textbf{Example images from our dataset.}
}
  \vspace{-4mm}
  \label{fig:visualizations}
\end{figure*}

\clearpage
\newpage

{\small
\newpage
\bibliographystyle{ieee_fullname}
\bibliography{egbib}
}